# Context-Enhanced Contrastive Search for Improved LLM Text Generation

Jaydip Sen
Department of Data Science
Praxis Business School
Kolkata, INDIA
email: jaydip.sen@acm.org

Rohit Pandey
Department of Data Science
Praxis Business School
Kolkata, INDIA
email: rohit.pandeyecon@gmail.com

Hetvi Waghela
Department of Data Science
Praxis Business School
Kolkata, INDIA
email: waghelah@acm.org

*Abstract*—Recently, Large Language Models (LLMs) have demonstrated remarkable advancements in Natural Language Processing (NLP). However, generating high-quality text that balances coherence, diversity, and relevance remains challenging. Traditional decoding methods, such as bean search and top-k sampling, often struggle with either repetitive or incoherent outputs, particularly in tasks that require long-form text generation. To address these limitations, the paper proposes a novel enhancement of the well-known Contrastive Search algorithm, Context-Enhanced Contrastive Search (CECS) with contextual calibration. The proposed scheme introduces several novelties including dynamic contextual importance weighting, multi-level Contrastive Search, and adaptive temperature control, to optimize the balance between fluency, creativity, and precision. The performance of CECS is evaluated using several standard metrics such as BLEU, ROUGE, and semantic similarity. Experimental results demonstrate significant improvements in both coherence and relevance of the generated texts by CECS outperforming the existing Contrastive Search techniques. The proposed algorithm has several potential applications in the real world including legal document drafting, customer service chatbots, and content marketing.

*Keywords*—Large Language Models (LLMs), Contrastive Search, Text Generation, Contextual Calibration, Coherence Optimization, Adaptive Decoding, Creative Content Generation, Natural Language Processing (NLP).

## I. Introduction

In recent years, Large Language Models (LLMs) have transformed the field of Natural Language Processing (NLP), delivering cutting-edge performance across numerous tasks, including text generation, summarization, machine translation, and question answering. Models such as OpenAI's GPT-3 [1], Google's BERT [2], and more recently PaLM [3], have greatly enhanced the capabilities of machines in understanding and generating human language. By leveraging deep neural network architectures and training on extensive datasets, LLMs have made significant strides in producing fluent and coherent text that closely resembles human communication.

Generating text from an LLM involves more than simply predicting the next word in a sequence according to its probability distribution. This step, known as decoding, plays a critical role in shaping the final output. Various decoding strategies have been proposed in the literature ranging from deterministic methods such as beam search, to stochastic methods like top-k and nucleus sampling. While the deterministic methods choose the highest probability token at each step, their stochastic counterparts introduce randomness to improve diversity in the generated output. These techniques have enabled LLMs to achieve impressive results. However, they have several limitations, particularly when applied to tasks that require a fine balance between coherence, diversity, and contextual accuracy [4].

For instance beam search, while effective at generating highly likely sequences, often results in repetitive or overly deterministic text, making the output less creative and sometimes monotonous. On the other hand, sampling-based methods, such as top-*k* and nucleus sampling, while introducing diversity, can generate incoherent or irrelevant content, especially when the model's predictions are uncertain or ambiguous. These shortcomings present a significant challenge, particularly for long-form text generation, where maintaining both coherence and relevance is crucial.

To resolve these issues, the field of LLM has witnessed the emergence of more sophisticated decoding strategies that attempt to balance the benefits of both deterministic and stochastic approaches. One such method that gained a high popularity is Contrastive Search [5]. Contrastive Search employs a contrastive scoring function that balances the need for coherence with the desire for diverse and creative outputs. The key idea behind this algorithm is to compare candidate tokens not only based on their probability of being the next token but also on how they contribute to the diversity of the generated sequence. This allows the decoder to produce text that is both fluent and non-repetitive, avoiding overly deterministic and incoherent outputs. While Contrastive Search has demonstrated promising results, further improvements are needed to optimize the generation process, especially in applications that demand high precision, coherence, and contextual relevance over long sequences [6].

The motivation for this work arises from the need to improve the quality of text generated by LLMs, particularly in scenarios where the generated content must be both contextually accurate and coherent over long sequences of text. The limitations of traditional decoding techniques, such as repetition and lack of diversity and fluency, become even more pronounced when generating text for tasks that require attention to detail. In these cases, the generated text must be accurate while maintaining coherent, relevant, and consistent with the input context. Classical decoding strategies most often struggle to meet these demands, particularly when the inputs are complex and ambiguous [7].

This paper proposes Context-Enhanced Contrastive Search (CECS), a novel enhancement to the Contrastive Search algorithm designed to address its limitations. The proposed method introduces several novel features designed to enhance LLM's ability to produce coherent, contextually appropriate, and diverse text for downstream applications.

The key contributions of this work are four-fold. *First*, the proposed method introduces a dynamic weighting mechanism that adjusts the significance of different sections of the input context depending on their relevance. This ensures that the most critical sections of the input receive more attention during the generation process, leading to more coherent and contextually appropriate outputs. The importance weighting is achieved through attention mechanisms that analyze the input in real time and dynamically adjust the weights assigned to each part of the context. *Second*, the proposed scheme extends the traditional Contrastive Search algorithm by introducing a multi-level search strategy that operates at different granularities. At the highest level, sentence-level coherence is ensured by first generating a rough outline of the text. This is followed by a more detailed phrase-level and word-level refinement, where the search algorithm focuses on improving the fluency and precision of the text at a finer level. This hierarchical approach allows the model to maintain both macro-level coherence and micro-level precision, resulting in outputs that are coherent and contextually relevant from start to finish. *Third*, a dynamic temperature control mechanism is introduced to adjust the randomness of the generation process based on the complexity of the input. In simpler, and more straightforward contexts, the temperature is kept low to ensure precise and deterministic outputs. In more complex or ambiguous contexts, the temperature is increased to allow for greater creativity and diversity in the generated text. This adaptation is guided by monitoring the entropy of the model's output distribution, ensuring that the balance between creativity and accuracy is maintained throughout the generation process. *Fourth and final*, a hybrid contrastive-divergence approach is incorporated in the scheme, where generated hypotheses are compared against negative samples to refine the output iteratively. By contrasting potential outputs with undesirable candidates, the model is steered away from incoherent or irrelevant paths, ensuring that the final output remains contextually appropriate and fluent.

The paper is structured as follows. An overview of recent advancements in text generation and decoding algorithms is presented in Section II. Section III details the proposed CECS algorithms, including its formulation and underlying principles. Section IV outlines the experimental results, comparing CECS with Contrastive Search on several metrics. Lastly, Section V concludes the paper highlighting some future works.

## II. RELATED WORK

Decoding strategies play a very critical role in shaping the effectiveness and quality of text generated by LLMs. As LLMs become increasingly sophisticated and expansive, the choice of decoding method directly impacts the fluency, coherence, and diversity of their output. The challenge lies in balancing these factors to avoid common issues such as repetitive or incoherent sequences. Researchers have proposed various decoding techniques to address these challenges. In this section, some of these schemes are discussed highlighting their contributions and limitations.

BehnamGhader et al. demonstrate how LLMs can effectively function as powerful text encoders and provide meaningful sentence embeddings that outperform traditional models like Sentence-BERT in various downstream tasks [8]. The authors introduce a framework for evaluating LLMs as text encoders and conduct extensive experiments to showcase their efficacy across different datasets and tasks. However, the work has limitations as it focuses on specific tasks and it relies on fine-tuned models which could restrict the generalizability of the findings across diverse applications.

Cai et al. introduce *Medusa*, a scheme to improve inference in LLMs by utilizing multiple parallel decoding heads [9]. The approach significantly reduces inference latency while possibly improving text-generating quality. However, its limitations include potential overhead in managing multiple decoding heads and the challenge of scaling this method efficiently for extremely large models.

Yan et al. introduce a speculative decoding approach that aims at accelerating text generation in LLMs by predicting multiple tokens in parallel [10]. The novelty of the scheme lies in its ability to make speculative predictions in parallel, which speeds up decoding without compromising on accuracy or fluency of the generated text. However, its dependence on accurate early token predictions is a serious limitation.

Li et al. present a framework that extends the context window of LLMs by employing parallel decoding, enabling efficient handling of longer sequences without sacrificing the quality of the generated text [11]. While the scheme maintains scalability and context comprehension, reduced coherence in the generated text is a major limitation.

Welleck et al. propose unlikelihood training, a method designed to reduce common issues in text generation, such as repetitions and incoherence, by explicitly penalizing undesired token sequences during training [12]. The core contribution of the approach is its success in mitigating repetitive outputs and improving text diversity across various natural language generation tasks. However, its limitation lies in the complexity of fine-tuning the unlikelihood loss, which may not generalize across all text generation use cases.

Zhu et al. propose an adaptive temperature sampling technique to enhance the performance of LLMs in code generation tasks by dynamically adjusting the temperature parameter based on the context and complexity of the task [13]. While the key contribution of the scheme is its ability to balance between deterministic (cold) and creative (hot) outputs, its limitation lies in the increased complexity of dynamically tuning the temperature for different text generation tasks.

Meister et al. propose a new decoding algorithm called locally typical sampling, which aims to generate text that is both fluent and natural [14]. The approach relies on producing tokens that are representative of the local context, rather than globally likely. The key contribution of this proposition is its ability to reduce common text generation issues such as incoherence or excessive diversity by ensuring that the generated tokens align closely with typical patterns observed in human language. However, the approach may not always promote creativity or variability.

Shi et al. present a new decoding method that incorporates a context-aware approach to minimize hallucinations in text generation from LLMs [15]. The primary contribution of this work is its ability to reduce the generation of factually incorrect or irrelevant information by ensuring that the generated text remains grounded in the provided context. However, issues like additional computational complexity and the need for precise context management pose serious limitations to the scheme for real-time.

Leviathan et al. propose a speculative decoding technique to accelerate inference in transformer-based models by generating multiple token predictions simultaneously [16]. This allows for faster text generation without a significant loss in quality. The key contribution of the approach is its ability to reduce latency in transformer models, making it suitable for real-time applications. However, the scheme suffers from a drawback that lies in its reliance on accurate early token predictions, as incorrect initial guesses can lead to inefficiencies or degraded quality.

Chen et al. present a speculative sampling technique that attempts to speed up text generation in LLMs by predicting multiple tokens in parallel and then refining the output to ensure quality [17]. The core contribution of the scheme is its ability to significantly reduce inference time while maintaining coherence and accuracy. However, the limitations include its reliance on accurate speculative predictions, which, if incorrect, may require costly adjustments and reduce efficiency in certain contexts.

Liu et al. present a benchmark specifically tailored to assess the performance of LLMs in tasks involving long-context generation tasks [18]. The primary contribution of this research is the creation of a comprehensive framework for assessing LLM's capabilities in handling long-range dependencies. However, the proposed approach may not fully capture all nuances of long-context generation across different languages and domains, and some models could still struggle with extreme cases of long-term coherence.

Despite significant progress in decoding algorithms for LLMs, there remain several gaps in the current state of the art. Approaches such as beam search, nucleus sampling, and top-$k$ sampling struggle to balance fluency, diversity, and coherence in generated text, often leading to repetitive outputs or hallucinations. While speculative decoding and parallel decoding introduced notable improvements in inference speed, they often rely on the assumption of accurate initial token predictions, which can degrade performance in more complex scenarios. Most importantly, the existing methods frequently fail to adapt dynamically to context changes during generation, which results in inconsistent or suboptimal outputs, especially in long-form texts. The current work seeks to address these limitations by introducing context awareness in the Contrastive Search, making it a more powerful decoding scheme. The proposed method adapts dynamically to the nuances of text generation, improving both efficiency and the quality of the output text without compromising on diversity or coherence. By building on Contrastive Search principles while incorporating context sensitivity and adaptive mechanisms, the proposed scheme aims to close the gap between static decoding techniques and the increasing dynamic requirements of modern LLM applications.

### III. CONTEXT-ENHANCED CONTRASTIVE SEARCH

The proposed Context-Enhanced Contrastive Search (CECS) framework aims to enhance the decoding process for LLMs by introducing a dynamic and context-sensitive approach to text generation. CECS adjusts key parameters such as temperature, token selection thresholds, and penalty mechanisms based on the evolving context of the generated text. This allows CECS to maintain a balance between fluency, coherence, and diversity throughout the decoding process, overcoming the limitations of static techniques like beam search and top-$k$ sampling. By incorporating dynamic temperature scaling, context-sensitive token selection, and an adaptive penalty system, CECS delivers more efficient and contextually relevant outputs, especially in tasks requiring long-range coherence and real-time interactions. An in-depth analysis of the mathematical framework is and the components of the CECS algorithm is provided in the following.

In defining the problem and developing the proposed solution framework, the following notations are used. Let:

$p(y_t|y_{1:t-1}, x)$ be the conditional probability of generating token $y_t$ given previous tokens $y_{1:t-1}$ and $x$.

$V$ be the vocabulary of the model.

$T$ be the total number of tokens to generate.

$P(y_t|y_{1:t-1})$ represent the probability distribution over $V$ for token $y_t$.

The objective of CECS is to produce a sequence of words $y_1, y_2 \ldots y_T$ such that it maximizes both diversity and coherence while maintaining efficiency in real-time scenarios.

#### A. Contrastive Search Framework

A decoding algorithm faces the challenge of balancing two aspects, fluency and diversity. Fluency ensures the generated text reads naturally and adheres to grammatical norms, maintaining relevance to the preceding context. Diversity, on the other hand, avoids repetition, promotes creativity, and allows the model to generate varied and non-deterministic outputs. Contrastive search aims to resolve this dilemma by dynamically weighing the importance of fluency and diversity during decoding. It selects tokens that strike a balance between maximizing probability (i.e., fluency) and maintaining diversity. The core of the Contrastive Search algorithm is a scoring function, as defined in (1).

$$S(y_t) = \lambda * \log p(y_t|y_{1:t-1}, x) - (1-\lambda) * D(y_t, y_{1:t-1}) \quad (1)$$

In (1), $\log p(y_t|y_{1:t-1}, x)$ is the log-probability of token $y_t$, $D(y_t, y_{1:t-1})$ is a diversity-promoting term that measures how distinct the token $y_t$ is from previous tokens $y_{1:t-1}$, and $\lambda$ is a balancing parameter between fluency and diversity.

#### B. Dynamic Temperature Scaling

Dynamic temperature scaling is a key component of the CECS framework that allows the model to dynamically adjust the sharpness or smoothness of its token probability distribution based on the context in which the text is being generated. In the context of text generation, temperature is a parameter that is utilized to adjust the degree of softness or sharpness in the probability distribution over possible next tokens. A low temperature (typically less than 1) concentrates the probability mass of the most likely tokens making the model more confident in its choice for the next token. On the other hand, a high temperature (typically greater than 1) spreads out the probability mass across a wider range of tokens and introduces more uncertainty and randomness in token selection. In the original Contrastive Search, a fixed temperature is applied throughout the decoding process. However, this leads to suboptimal text generation as different parts of a text sequence require different levels of creativity or determinism. For instance, while generating a technical document, most of the text should be coherent and logical (i.e., a low temperature), but certain sections might benefit from more diversity and creativity (i.e., high temperature). To improve upon the static temperature typically used in decoding strategies, CECS introduces a novel concept called *temperature scaling*. At each decoding step *t*, the temperature $\tau_t$ is dynamically adjusted based on the complexity of the generated text so far (also referred to as the context complexity). The idea is to introduce a more flexible approach where simpler contexts call for lower temperatures favoring fluency. On the other hand, more complex or diverse contexts call for higher temperatures for higher diversity. The temperature parameter is dynamically adjusted according to the present state of the generation process using (2)

$$\tau_t = \tau_0 * (1 + \alpha * f_{context}(y_{1:t-1})) \qquad (2)$$

In (2), $\tau_0$ is the initial temperature set by the user or based on the nature of the task, α is a scaling factor that controls how sensitive the temperature adjustment is to the context complexity, $f_{context}(y_{1:t-1}))$ is a function that measures the complexity or uncertainty in the current context $y_{1:t-1}$.

*Context complexity* is a dynamic metric that evaluates how varied or predictable the generated text is up to the current point. This can be computed in three different ways.

*Token diversity*: This method involves measuring the diversity of the tokens in the current context using the lexical diversity metrics.

*Entropy of the token probability distribution*: A higher entropy indicates that the model is less confident about the next token, indicating a more complex and uncertain context.

*Length of structural complexity*: Longer or structurally more complex sentences require higher temperatures to introduce more creative transitions.

The dynamic temperature scaling allows the model to reduce temperature in predictable contexts favoring more deterministic outputs, and increase it when more diversity or creativity is required.

As an *illustrative example* of dynamic temperature scaling, let's consider generating a narrative text. At the beginning of a sentence, where more deterministic, fluent choices are necessary to ensure grammatical correctness, the model might apply a low temperature, say, $\tau_t = 0.7$. Suppose, the *input text* is: "The sun was setting, and", the model's *confidence is high* (i.e., predictable context), and the resulting *temperature is low* (favoring fluency). Later, as the sentence unfolds into more open-ended or creative territory, the context complexity increases (e.g., the model is faced with many possible ways to complete the sentence). Suppose, the current input text is: "The sun was setting, and the sky turned...", the model's confidence is now low (i.e., many possible continuations: red, orange, dark, stormy, etc.), and hence, the resulting temperature increases dynamically to $\tau_t = 1.5$ (say), favoring diverse and creative completions. This adaptive mechanism ensures that the model generates rich and varied outputs when appropriate while maintaining coherence and grammatical integrity when necessary.

*C. Context-Sensitive Token Selection*

Context-sensitive token selection is a crucial feature of CECS that allows it to dynamically choose tokens based on both the local (i.e., immediate sequence) and global (i.e., overall text context) considerations. This enhances the quality of the generated text by achieving a balance in semantic coherence, relevance to the task, and diversity in language. Context-sensitive token selection in CECS addresses the limitations of the original Contrastive Search by making token decisions that are contextually appropriate. The original Contrastive Search [5] introduced an idea of contrasting between high-probability and low-probability tokens within the same distribution to avoid the limitations of conventional search methods such as greedy or beam search. In CECS, this framework is enhanced to make token selection context-sensitive. In CECS, two types of tokens are considered. The *high-confidence tokens* are the tokens to which the model assigns a high probability based on the immediate context. In regular decoding, these tokens would typically be selected, as they are the most likely candidates to maintain fluency and grammatical correctness. The *low-confidence tokens*, on the other hand, are less likely candidates that might introduce more diversity into the generated text. However, if chosen indiscriminately, they can cause incoherence and irrelevance in the generated text. CECS leverages both categories of tokens through contrastive scoring and ensures that only tokens contributing to overall fluency, coherence, and diversity are selected.

CECS incorporates a mechanism called *contextual calibration* that ensures that the decoder dynamically adapts to the evolving complexity of the text. At each token selection step, the model looks at the previous tokens and the global task at hand. The previous tokens are used to track the ongoing sentence or paragraph structure to ensure that the new tokens fit the grammatical and semantic flow. The model also takes into account the global task at hand, such as the thematic or domain-specific focus.

CECS introduces a context-sensitive token selection method in which the set of candidate token $C_t$ is dynamically determined based on the context $y_{1:t-1}$, at each decoding step $t$. Instead of fixed thresholds for sampling, CECS uses a dynamic selection criterion specified in (3).

$$C_t = \{y \in V : p(y_t|y_{1:t-1}, x) \geq \beta_t * \max_{y' \in V} p(y'|y_{1:t}, x)\} \qquad (3)$$

In (3), $\beta_t$ is a context-dependent threshold that adjusts dynamically at each step and ensures that the model adapts to the level of certainty or uncertainty in its predictions. $\beta_t$ is given by (4)

$$\beta_t = \beta_0 * g_{context}(y_{1:t-1}) \qquad (4)$$

In (4), $\beta_0$ is an initial baseline threshold, and $g_{context}$ is a function that adjusts the threshold based on the entropy or variance in the token distribution. The function $g_{context}$ reflects how confident or uncertain the model is in its predictions. This mechanism ensures that when the model is confident (i.e., its entropy is low), $g_{context}$ reduces the threshold $\beta_t$, and hence, fewer candidates are considered. In uncertain contexts, when the model's confidence is low (i.e., its entropy is high), $g_{context}$ increases $\beta_t$, expanding the set of candidate tokens. This context-sensitive adaptation enables more robust text generation, especially in complex or ambiguous scenarios, while maintaining coherence in more straightforward contexts.

For a token $y$ to be included in the candidate set $C_t$, its probability must be at least $\beta_t$ times the maximum probability in the current step. In simpler terms, only tokens whose probabilities are relatively close to the highest-probability token are considered.

*D. Adaptive Penalty Mechanism*

To reduce repetition and ensure more coherent outputs, CECS includes an adaptive penalty mechanism. This mechanism dynamically adjusts the degree of penalty for specific tokens, making it context-aware so that issues such as overgeneration, repetition, and low creativity in generated text are mitigated. At each decoding step $t$, when selecting the next token, the adaptive penalty mechanism adjusts the probability distribution of tokens as determined by their occurrence in the present context. The penalized probability $p_{pen}(y_t|y_{1:t-1}, x)$ for a token $y_t$ is computed using (5)

$$p_{pen}(y_t|y_{1:t-1}, x) = p(y_t|y_{1:t-1}, x) * \lambda_t(y_t) \quad (5)$$

In (5), $p(y_t|y_{1:t-1}, x)$ is the original probability of the token $y_t$ at step $t$, given the context $y_{1:t-1}$ and input $x$. The penalty factor $\lambda_t(y_t)$ adjusts the probability of $y_t$ based on its prior occurrence and contextual relevance. The penalty factor $\lambda_t(y_t)$ is determined using (6)

$$\lambda_t(y_t) = 1 - \alpha_t * f_{penalty}(y_t, y_{1:t-1}) \quad (6)$$

In (6), $\alpha_t$ is a scaling factor that controls the strength of the penalty at step $t$, $f_{penalty}(y_t, y_{1:t-1})$ is a penalty function that computes the penalty for the token $y_t$ based on its prior occurrence in the context $y_{1:t-1}$. The details of the penalty are outlined in the following. The computation of $\alpha_t$ is done later in this section.

The *penalty function* measures the extent to which a token has been overused or is inappropriate based on the current context. This function can take different forms depending on the desired behavior of the decoder. It it typically reflects two key factors: repetition penalty and contextual appropriateness. These two factors are discussed in the following.

*Repetition Penalty*: This component discourages the repeated use of tokens within a short span. If a token has already appeared in the sequence $y_{1:t-1}$, it receives a higher penalty. The penalty of repetition is computed using (7).

$$f_{penalty,rep}(y_t, y_{1:t-1}) = \frac{count(y_t, y_{1:t-1})}{t} \quad (7)$$

In (7), $count(y_t, y_{1:t-1})$ is the number of times $y_t$ has appeared in the sequence $y_{1:t-1}$, $t$ is the current decoding step. Tokens that have been used frequently in the recent past receive a higher penalty, and their probability of being selected again is reduced.

*Contextual Appropriateness*: The adaptive penalty mechanism of CECS also considers the contextual relevance of tokens. Some tokens may be inappropriate or less relevant based on the semantic context, and these tokens receive a higher penalty. This can be based on a measure of semantic similarity or coherence, adjusting the probability of tokens that do not align well with the current text. The penalty for contextual irrelevance is computed using (8).

$$f_{penalty,context}(y_t, y_{1:t-1}) = 1 - \cos(E(y_t), C(y_{1:t-1})) \quad (8)$$

In (8), $E(y_t)$ represents the embeddings of the candidate token $E(y_t)$, $C(y_{1:t-1})$ is the contextual embedding of $y_{1:t-1}$, and $\cos(.,.)$ measures the cosine similarity between $E(y_t)$ and $C(y_{1:t-1})$. Incorrect token selection is avoided by giving higher penalties to tokens that are semantically distant from the context.

*Adaptive Scaling of the Penalty Factor*: One of the most critical features of CECS is the dynamic scaling of the penalty factor $\alpha_t$, which controls the degree of penalty applied at each step. This factor is context-sensitive and changes based on the model's level of confidence and the complexity of the current context. The scaling factor is determined by a function that evaluates the entropy or variance of the token probability distribution using (9).

$$\alpha_t = \alpha_0 * h_{context}(y_{1:t-1}) \quad (9)$$

In (9), $\alpha_t$ is the scaling factor at step $t$, $\alpha_0$ is the baseline penalty strength, $h_{context}(y_{1:t-1})$ is a function that evaluates the complexity of the current context, typically through the entropy or variance of the token probability distribution. When the token probabilities are spread out indicating uncertainty or complexity, $\alpha_t$ increases, and stronger penalties are applied to prevent degenerate or repetitive text generation. When the model is confident and the token probabilities are concentrated on a few high-probability tokens, $\alpha_t$ decreased so that a more conservative selection of high-probability tokens is made to maintain coherence in the generated text.

```
Algorithm: Adaptive Contrastive Search (ACS)

Input:
    - Model: LLM (Large Language Model)
    - Input x: Input text
    - T: Maximum number of decoding steps
    - λ: Weighting factor for fluency-diversity tradeoff
    - τ_0: Initial temperature
    - γ: Penalty factor for token repetition

Output:
    - Generated sequence y = {y_1, y_2, ..., y_T}

1. Initialize:
    - y_0 ← [START]    // Initialize sequence with start token
    - τ ← τ_0          // Set initial temperature
    - Set current context C ← {y_0}

2. For each time step t from 1 to T:
3.      Compute the conditional probability distribution:
            - p(y_t | C) = Model(C)   // Use the LLM to get token probabilities

4.      Apply adaptive penalty for repeated tokens:
            - For each token y in the vocabulary V:
                - p_adjusted(y_t) ← p(y_t | C) * (1 - γ * repetition_penalty(y_t, C))

5.      Dynamically adjust temperature based on context:
            - τ_t ← τ_0 * (1 + α * context_complexity(C))   // Adjust temperature based on cont

6.      Select candidate tokens based on context-sensitive filtering:
            - C_t ← {y ∈ V : p_adjusted(y_t) ≥ β_t * max(p_adjusted(y_t))}
            // β_t is context-dependent threshold

7.      Score candidate tokens using contrastive search objective:
            - For each y in C_t:
                - S(y_t) ← λ * log(p_adjusted(y_t)) - (1 - λ) * diversity(y_t, C)

8.      Select token y_t that maximizes the contrastive search score:
            - y_t ← argmax(S(y_t))

9.      Update the context:
            - C ← C ∪ {y_t}

10.     If y_t is the end token or maximum steps reached:
            - Break

11. Return generated sequence y = {y_1, y_2, ..., y_T}

End Algorithm
```

Fig. 1. The pseudocode for the CECS text decoding algorithm

### E. Decoding Procedure

The decoding procedure of CECS introduces a dynamic and context-aware mechanism to generate high-quality text from LLMs. The process begins by initializing the sequence with a start token and progressively builds the output by selecting tokens one by one. At each decoding step, CECS first computes the conditional probabilities of all possible next tokens based on the current context. A penalty is applied to discourage repeated tokens, and the temperature is adjusted in response to the complexity of the context. This enables the decoder to have finer control over the fluency and diversity of the output. Candidate tokens are filtered using a context-sensitive threshold, and the Contrastive Search objective is applied to balance between selecting tokens that maximize fluency while preserving diversity. Fig 1 depicts pseudocode for the CECS decoding process.

IV. EXPERIMENTAL RESULTS

To compare the performance of CECS with the original Contrastive Search algorithm, the experimental framework outlined in [5] is followed. The tasks for evaluation, including open-ended text generation, document summarization, and machine translation, are selected to measure fluency, coherence, and diversity in the generated outputs.

Automatic metrics such as diversity [5], MAUVE [19], and coherence [5] are used to quantify the quality and diversity of the text used in evaluation across various language models (e.g., GPT-2 [20], GPT-Neo [21], and OPT [22]), varying context and generation length to compare the two algorithms under diverse conditions.

*Diversity* represents an overall measure of the repetition in the generated text at different levels of *n*-grams (i.e., a contiguous sequence of *n* item tokens), with *n* typically set of values like {2, 3, 4}. This metric evaluates the tendency of the model to avoid repetitive patterns by analyzing how often distinct *n*-grams are generated in the text. The diversity metric (*D*) evaluates the percentage of unique *n*-grams within the generated sequence relative to the total number of *n*-grams, and it is computed using (10).

$$D = 1 - \frac{Count\ of\ repeated\ n\_grams}{Total\ count\ of\ n\_grams} \quad (10)$$

*Measuring Alignment Using Variational Embeddings* (MAUVE) is a metric designed to evaluate how closely the distribution of machine-generated text matches that of human-written text. It compares the token distributions of both to assess how neural and human-generated text match. MAUVE calculates a divergence score between the two distributions using a statistical method that captures differences in quality and fluency. A higher MAUVE score indicates that the model-generated text is more aligned with human-written text in terms of structure, diversity, and fluency.

*Coherence* measures the semantic connection and consistency between the generated text and the initial prefix text provided to a language model [5]. It evaluates how well the language model maintains the logical flow and meaning throughout the text generation process. It is computed as the average log-likelihood of each token in the generated text, given the context of all prior tokens, using a pre-trained language model such as OPT. This determines how well the generated text aligns semantically with the prompt. Coherence (*C*) is computed using (11).

$$C(\hat{x}, x) = \frac{1}{|\hat{x}|} \sum_{i=1}^{|\hat{x}|} \log(p_M(\hat{x}_i | x, \hat{x}_{<i})) \quad (11)$$

In (11), $\hat{x}_i$ is the generated text, $x$ is the prefix text, $p_M(\hat{x}_i | x, \hat{x}_{<i})$ is the probability of the token $\hat{x}_i$ given both the prefix $x$ and the previously generated tokens $\hat{x}_{<i}$. The average log-likelihood is taken over the length of the generated text $|\hat{x}|$. The coherence metric used in the current work as given by (10) slightly different from the formulation in [5]. The difference here is that the probability distribution is explicitly conditioned on both the prefix $x$ and the full preceding generated tokens $\hat{x}_{<i}$, which slightly shifts the focus to emphasize the joint contribution of both the prefix and the past sequence in generating coherent content.

*A. Open-Ended Text Generation*

Following the approach presented in [5], to address potential inaccuracies in the coherence measurement due to the inductive biases of the model *M*, the coherence scores are presented using different scales of the OPT model, including 125 M, 2.7M, and 13B parameters. The average length of the generated text, *gen-length*, is also reported for the two decoding methods to assess their text generation efficiency.

TABLE I. TEXT GENERATION PERFORMANCE ON THE WEBTEXT DATASET

| Decoding Algo | | Contrastive Search | CECS |
|---|---|---|---|
| Diversity (%) | | 92.54 | 94.37 |
| MAUVE (%) | | 87.26 | 90.28 |
| Gen-Length | | 140.72 | 141.87 |
| Coherence | OPT-125 Million | -1.93 | -1.42 |
| | OPT-2.7 Billion | -1.52 | -1.22 |
| | Opt-13 Billion | -1.56 | -1.28 |

Following the experimental setup for Contrastive Search in [5], the large GPT-2 model is employed to generate text sequences based on the initial 40 tokens from documents in the reserved portion of the WebText dataset [20]. The generation process is terminated when either an end-of-document token is encountered or a maximum length of 200 tokens is reached. For Contrastive Search, the parameters $k = 5$ and $\lambda = 0.4$ are used. The value of *k* refers to the number of candidate tokens considered during the decoding process, and it controls the trade-off between diversity and fluency in the generated text. The parameter λ controls how much emphasis is placed on fluency versus diversity. A higher λ prioritizes fluency, whereas a lower value λ prioritizes diversity.

Table I highlights several improvements CECS brings forth in comparison to Contrastive Search. A higher diversity value for CECS indicates CECS reduces token repetition and generates a broader range of text, making it less likely to repeat phrases or structures. The higher MAUVE score of CECS suggests that the text generated by CECS is more similar to human-written text. The small increase in gen-length indicates that CECS might allow for more extended or elaborate content generation. Finally, across all three scales of the OPT model, CECS is found to deliver better coherent scores. This signifies that CECS is capable of maintaining better semantic consistency with the input text.

*B. Document Summarization*

For the document summarization task, based on the setup described in [5], the XSum dataset [23], consisting of news articles from the BBC along with corresponding one-sentence summaries, is employed for testing. Text summarization tasks are performed using OPT models with different scales (125 M to 2.7 B parameters) applying both Contrastive Search and CECS. As in open-ended text generation tasks, for summarization too, the parameters $k = 5$ and $\lambda = 0.4$ are used for Contrastive Search. Following the experiments in [5], two scenarios are explored: (i) *zero-shot learning*, and (ii) *in-*

*context learning*. In the zero-shot setting, the model is prompted with a document and it generates a summary autoregressively. For in-context learning, one or two article-summary pairs are provided as examples to guide the model during summary generation [1].

Document summarization is evaluated using the *recall-oriented understudy for gisting evaluation* (ROUGE) metrics [24]. The ROUGE score is a set of metrics that compares summaries or translations with reference texts (usually human-written). In comparing the document summarization performance of Contrastive Search with CECS, ROUGE-1, ROUGE-2, and ROUGE-L scores are used.

ROUGE-1 measures the overlaps of single words (unigrams) between the generated text and the reference text. It evaluates how much of the reference content is covered in the generated text in terms of individual word matching, and computed using (12).

$$ROUGE_1 = \frac{No\ of\ overlapping\ unigrams}{Total\ no\ of\ unigrams\ in\ the\ ref\ text} \quad (12)$$

ROUGE-2 measures the overlap of two-word sequences (bigrams) between the generated and reference texts. It captures the coherence of phrases, as it looks at how well consecutive word pairs in the reference are represented in the generated text. The computation of ROUGE-2 is similar to that of ROGUE-1.

ROUGE-L measures the *longest common subsequence* (LCS) between the generated and reference texts. It evaluates both content coverage and the fluency of the generated text by considering the longest matching sequence of words. The computation of ROUGE-L involves finding the LCS between the reference and generated text and normalizing it by the length of the reference text using (13).

$$ROUGE_L = \frac{LCS\ between\ generated\ and\ reference\ text}{Length\ of\ reference\ text} \quad (13)$$

ROUGE-L is more useful for evaluating sentence-level coherence and ordering than ROUGE-1 and ROUGE-2.

TABLE II. SUMMARIZATION PERFORMANCE ON THE XSUM DATASET

| Language Models | Shots | Contrastive Search | | | CECS | | |
|---|---|---|---|---|---|---|---|
| | | R-1 | R-2 | R-L | R-1 | R-2 | R-L |
| OPT-125 Million | Zero | **12.68** | 1.75 | 9.59 | 9.56 | **1.96** | **11.12** |
| | One | 15.86 | 1.96 | 12.03 | **16.64** | **2.84** | **14.32** |
| | Two | 18.04 | 2.63 | 13.89 | **20.34** | **3.18** | **15.65** |
| OPT-350 Million | Zero | 1.11 | **0.16** | 0.86 | **3.12** | 0.12 | **2.17** |
| | One | 17.30 | 2.67 | 13.27 | **20.59** | **4.02** | **18.36** |
| | Two | 18.84 | 3.01 | 14.48 | **22.41** | **5.07** | **18.58** |
| OPT-1.3 Billion | Zero | 16.76 | **3.17** | 12.64 | **18.73** | 1.12 | **14.34** |
| | One | 25.36 | 6.57 | 19.76 | **29.62** | **9.42** | **25.83** |
| | Two | 27.31 | 7.54 | 21.16 | **33.72** | **12.36** | **27.84** |
| OPT-2.7 Billion | Zero | 4.95 | **1.03** | 3.81 | **6.83** | 0.08 | **6.53** |
| | One | 27.77 | 8.22 | 21.77 | **31.68** | **12.47** | **26.53** |
| | Two | 29.02 | 9.09 | 23.07 | **36.72** | **14.45** | **31.05** |

It is observed from Table II that under the zero-shot setting, the performance of the two methods exhibits fluctuations across different language models. In Table II, R-1, R-2, and R-L denotes ROUGE-1, ROUGE-2, and ROUGE-L scores, respectively. In zero-shot learning, the model relies entirely on its pre-existing knowledge gained during training. This makes it sensitive to the type of tasks or inputs given, as the model has to generalize purely based on the input it has seen during training. While larger models, with more parameters, tend to capture more complex patterns and dependencies in language, it doesn't always guarantee consistent performance across tasks. This can happen due to possible model overfitting and may make models misinterpret certain types of input. Another reason for the inconsistent performance of the algorithms for zero-shot settings is *inductive bias*. Inductive bias refers to the assumptions a model makes to generalize from the data it has seen to unseen examples. Smaller models might prioritize simple patterns, leading to more conservative outputs, while larger models may attempt more complex generalizations, which can sometimes result in better performance, but also with greater instability in outcomes depending on the input. This explains the fluctuations in performance of the two decoding algorithms across the language models in Table II. However, for one and two in-context learning scenarios, in the absence of any inductive bias in the models, CECS outperforms Contrastive Search yielding higher ROUGE scores consistently for document summarization tasks.

*C. Machine Translation*

Finally, experiments on the machine translation task are conducted. For machine translation tasks, two metrics BLEU and COMET are used.

*Bilingual Evaluation Understudy* (BLEU) is a metric for evaluating the quality of machine-generated translations compared to human translations [25]. It evaluates translations by comparing *n*-grams in the machine-translated text with those in one or more reference translations. BLEU calculates precision by comparing *n*-grams of the machine-generated translation with *n*-grams of the reference translations. It checks how many *n*-grams in the machine translation appear in the reference translations. To present very short machine translations that may artificially achieve high precision, the computation of BLEU includes a brevity penalty. This penalty is applied when the generated translation is shorter than the reference translation. The BLEU score is computed as a weighted geometric mean of the *n*-gram precisions, along with the brevity penalty using (14).

$$BLEU = BP * \exp(\sum_{n=1}^{N} w_n * \log(p_n)) \quad (14)$$

In (14), BP is the *brevity penalty*, $p_n$ is the n-gram precision, i.e., the number of *n*-grams that match between machine translation and references, and $w_n$ is the weight assigned to n-grams.

The brevity penalty (BP) is derived using (15).

$$BP = \begin{cases} 1 & if\ c > r \\ \exp(1 - r/c) & if\ c \leq r \end{cases} \quad (15)$$

In (15), *c* is the length of the generated text, and *r* is the length of the reference. BLEU score lies between 0 to 1.

*Cross-lingual Optimized Metric for Evaluation of Translation* (COMET) score is a neural framework designed to evaluate the quality of machine-translated text [26]. It utilizes pre-trained multilingual models, such as BERT [27] and XLM-R [28] to capture both semantic and linguistic aspects of translations. COMET employs a transformer-based neural architecture that uses pre-trained language models, fine-tuned on bilingual data. COMET models are trained to predict human judgments of translation quality by minimizing the error between its prediction and human-assigned scores. The training process involves human evaluations, where a dataset with human judgment on translations is used as ground truth. The model takes as input: (i) the source sentence, (ii) the machine-generated translation (hypothesis), and (iii) an optional human reference translation, depending on the type of COMET metric being used. It then compares the machine-generated translation with the human reference and generates a score indicating the quality of the translation. The score predicts how well the translation reflects meaning and fluency compared to human references. COMET is found to correlate better with human judgments compared to traditional metrics like BLEU, as it takes into account deeper contextual and semantic information [29].

Machine translation tasks are conducted using the IWSLT14 De-En dataset [30] in a similar line as presented in [5]. The same OPT models are tested with Contrastive Search ($k = 5$ and $\lambda = 0.4$) and CECS. The results are presented in Table III. It is observed that OPT-125 M and OPT-350 M models, i.e., smaller language models, do not achieve good BLEU scores for both algorithms. However, as the number of model parameters increases, for models of 1.3 B parameters or more, a significant improvement in performance is observed. This indicates the emergence of capabilities in larger models. The performance of CECS is found to be consistently superior to Contrastive Search across large language models, as evidenced by both BLEU and COMET scores. This indicates that CECS provides more accurate and semantically meaningful outputs in machine translation tasks.

TABLE III. MACHINE TRANSLATION RESULTS ON IWSLT14 DE-EN DATA

| Language Models | Shots | Contrastive Search | | CECS | |
|---|---|---|---|---|---|
| | | *BLEU* | *COMET* | *BLUE* | *COMET* |
| OPT-125 Million | One | 0.05 | -1.18 | **0.10** | **-1.10** |
| | Few | 0.05 | -1.38 | **0.12** | **-1.25** |
| OPT-350 Million | One | 0.00 | -1.30 | **0.08** | **-1.15** |
| | Few | 0.05 | -1.41 | **0.10** | **-1.20** |
| OPT-1.3 Billion | One | 7.10 | -0.41 | **7.80** | **-0.30** |
| | Few | 8.39 | -0.29 | **9.00** | **-0.15** |
| OPT-2.7 Billion | One | 12.98 | 0.04 | **13.50** | **0.08** |
| | Few | 13.52 | 0.05 | **14.00** | **0.10** |

*D. MAUVE-Coherence Tradeoff with K*

The study of the MAUVE-Coherence trade-off with varying values of *k* is crucial for understanding and optimizing the performance of text generation algorithms such as Contrastive Search and CECS. This trade-off sheds light on how well an algorithm balances human likeness (i.e., MAUVE) and semantic coherence as the number of candidate words (*k*) increases. MAUVE measures how closely the distribution of the generated text aligns with that of human-written text, indicating naturalness or fluency. On the other hand, coherence assesses the logical flow and semantic consistency of the text within the given context. This implies that a high coherence score ensures that the generated text follows the narrative or ideas set by the input. However, maximizing both metrics simultaneously is challenging. Increasing *k* improves diversity, as the model explores a wider range of candidate outputs. However, high diversity can lead to lower coherence since the model might explore more semantically divergent paths, which don't align with the context. A lower *k* value might improve coherence by restricting the generation to fewer, more contextually appropriate outputs, but at the expense of diversity and human likeness.

Understanding the trade-off helps in identifying the sweet spot where human likeness and coherence are maximized. In some tasks like creative writing or story generation, human likeness (MAUVE) may be more critical, allowing the text to flow naturally, even at the expense of strict coherence. In tasks such as summarization, translation, or factual text generation, coherence is paramount because the generated text must remain aligned with the input content.

The MAUVE-Coherence trade-off study serves as empirical evidence to guide algorithmic improvements. A more advanced algorithm is expected to offer a better balance of diversity and coherence. Following the experimental framework in [5], the MAUVE-Coherence trade-off as a function of *k* is studied for both Contrastive Search and CECS. The results are presented in Table IV. The results indicate that CECS provides a better balance between coherence and naturalness for a wider range of *k* values. Hence, CECS is more likely to maintain quality in text generation as the search becomes more diverse (i.e., *k* becomes larger). Fig 2 depicts Coherence vs MAUVE scores for Contrastive Search and CECS. The plot reinforces the idea that CECS is a more effective method for generating coherent and human-like text compared to Contrastive Search, particularly as the *k* varies.

TABLE IV. COHERENCE AND MAUVE SCORES FOR DIFFERENT K VALUES

| | Contrastive Search | | CECS | |
|---|---|---|---|---|
| *K* | *Coherence* | *MAUVE* | *Coherence* | *MAUVE* |
| 2 | -1.20 | 78.00 | -1.00 | 80.00 |
| 3 | -1.35 | 81.00 | -1.10 | 84.00 |
| 4 | -1.45 | 84.00 | -1.20 | 88.00 |
| 5 | -1.50 | 87.00 | -1.25 | 90.00 |
| 6 | -1.60 | 86.00 | -1.35 | 89.00 |
| 7 | -1.65 | 85.00 | -1.45 | 88.50 |
| 8 | -1.75 | 84.50 | -1.55 | 87.00 |
| 9 | -1.80 | 83.00 | -1.65 | 85.50 |
| 10 | -1.90 | 80.00 | -1.75 | 82.00 |

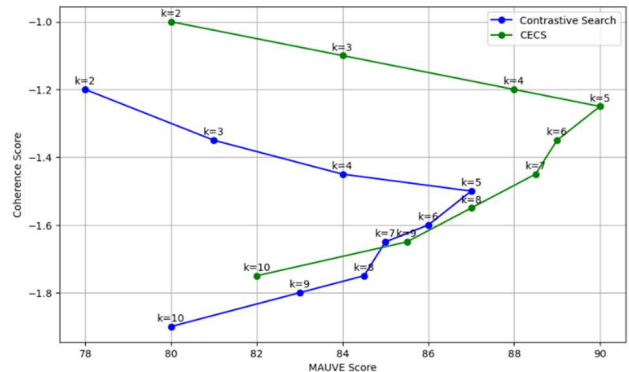

Fig. 2. MAUVE-Coherence trade-off for different values of *k*

V. CONCLUSION

In this paper, Context-Enhanced Contrastive Search (CECS) is introduced as a novel text generation algorithm that significantly enhances the coherence and quality of generated text compared to traditional Contrastive Search. The findings highlight the potential of CECS as a superior alternative for text generation tasks, particularly in applications requiring high-quality and coherent text.

A primary contribution of this work is the development of a comprehensive evaluation framework that allows for a systematic comparison between Contrastive Search and CECS. CECS is found to yield superior coherence scores, indicating that more logically structured and contextually relevant outputs are produced by it. This enhanced coherence is attributed to the adaptive nature of CECS, which dynamically adjusts its generation strategy based on input context and previous outputs. The analysis of MAUVE scores further highlights the effectiveness of CECS in generating text that closely aligns with human-written content.

Moreover, the relationship between the parameters of the algorithms, particularly the value of *k* in both Contrastive Search and CECS, is explored. The results reveal a trade-off between coherence and MAUVE scores, indicating that careful tuning of these parameters is necessary for optimal performance. Higher values of *k* generally enhance coherence, although their impact on MAUVE scores varies.

The comparative analysis of Contrastive Search and CECS reveals distinct behavioral patterns in text generation. While strengths in simpler generation tasks are demonstrated by Contrastive Search, higher-quality outputs are consistently delivered by CECS in more complex scenarios. This observation suggests that the adaptive mechanism of CECS is particularly beneficial in environments where context is dynamic and coherence is crucial. The significance of this study extends beyond theoretical contributions, as practical implications for various fields are identified. In NLP applications such as chatbots, content generation, and summarization tools stand to benefit from the advancements introduced by CECS. By enhancing the quality and coherence of generated text, improvements in user experiences and the overall effectiveness of NLP applications are anticipated.

Despite these positive findings, the limitations of this study must be mentioned. While comprehensive, the evaluation metrics may not capture every aspect of text quality. Future research could incorporate human evaluations to complement the automated metrics for a more holistic assessment of the generated text. Moreover, the scope of the study is limited to specific datasets and language models. Further experiments across a broader range of contexts and languages could enhance the generalizability of the findings.